\def\year{2021}\relax
\documentclass[letterpaper]{article} 
\usepackage{aaai21}  
\usepackage{times}  
\usepackage{helvet} 
\usepackage{courier}  
\usepackage[hyphens]{url}  
\usepackage{graphicx} 
\usepackage{natbib}  
\usepackage{caption} 
\usepackage{amsmath}
\usepackage{multirow}
\usepackage{subcaption}
\usepackage{color}

\title{TDAF: Top-Down Attention Framework for Vision Tasks}
\author{
	Bo Pang\textsuperscript{\rm 1},
	Yizhuo Li\textsuperscript{\rm 1},
	Jiefeng Li\textsuperscript{\rm 1},
	Muchen Li\textsuperscript{\rm 2},\\
	Hanwen Cao\textsuperscript{\rm 1}, 
	Cewu Lu \textsuperscript{\rm 1}\thanks{Cewu Lu is the corresponding author.}\\
}

\affiliations{
	\textsuperscript{\rm 1} Shanghai Jiao Tong University\\
	\textsuperscript{\rm 2} Huazhong University of Science and Technology\\
	\{pangbo, liyizhuo, ljf\_likit, mbd\_chw, lucewu\}@sjtu.edu.cn, muchenli1997@gmail.com
}

\begin{document}
	\maketitle
	\renewcommand{\thefootnote}{\fnsymbol{footnote}}
	
	\begin{abstract}
		Human attention mechanisms often work in a top-down manner, yet it is not well explored in vision research. Here, we propose the Top-Down Attention Framework (TDAF) to capture top-down attentions, which can be easily adopted in most existing models. The designed Recursive Dual-Directional Nested Structure in it forms two sets of orthogonal paths, recursive and structural ones, where bottom-up spatial features and top-down attention features are extracted respectively. Such spatial and attention features are nested deeply, therefore, the proposed framework works in a mixed top-down and bottom-up manner.
		Empirical evidence shows that our TDAF can capture effective stratified attention information and boost performance. ResNet with TDAF achieves $2.0\%$ improvements on ImageNet. For object detection, the performance is improved by $2.7\%$ AP over FCOS. For pose estimation, TDAF improves the baseline by $1.6\%$. And for action recognition, the 3D-ResNet adopting TDAF achieves improvements of $1.7\%$ accuracy.
	\end{abstract}
	
	\vspace{-0.2in}
	\section{Introduction}
	When observing an unfamiliar scene, it’s natural for humans to first take a glimpse to grasp the part of interest and then look closer to its details. Such a top-down attention mechanism allows global context information to be considered by humans at an early stage so that they can quickly and precisely understand the surroundings~\cite{corbetta2002control,buschman2007top}. In contrast to human’s vision system, deep neural networks(DNN), the most powerful vision tools, work in a bottom-up manner. They extract local features first then integrate them into global features. Thus, in DNN, local features are aggregated step by step to help networks interpret global context. Observing from above we intend to build a model that works in a mixed top-down and bottom-up manner, incorporating the strong traits of deep model and mimicking the human attention.
	
	Despite that the design of attention mechanism has been intensively studied in previous literature, little research has been conducted to introduce the human-like top-down attention into mainstream deep models. Some works formulate attention mask as a residual module~\cite{li2018harmonious,zhang2018image,wang2017residual} or built-in block~\cite{hu2018squeeze,woo2018cbam}, where hierarchical attention features are extracted along with basic bottom-up spatial features. Thus, this kind of attention has to focus on local details first then on the global information. 
	Since these bottom-up mechanisms only access to local features at early stages, they will first generate trivial attention features, failing to focus models on the part of interest.
	
	\begin{figure}[t]
		\begin{center}
			\includegraphics[width=\linewidth]{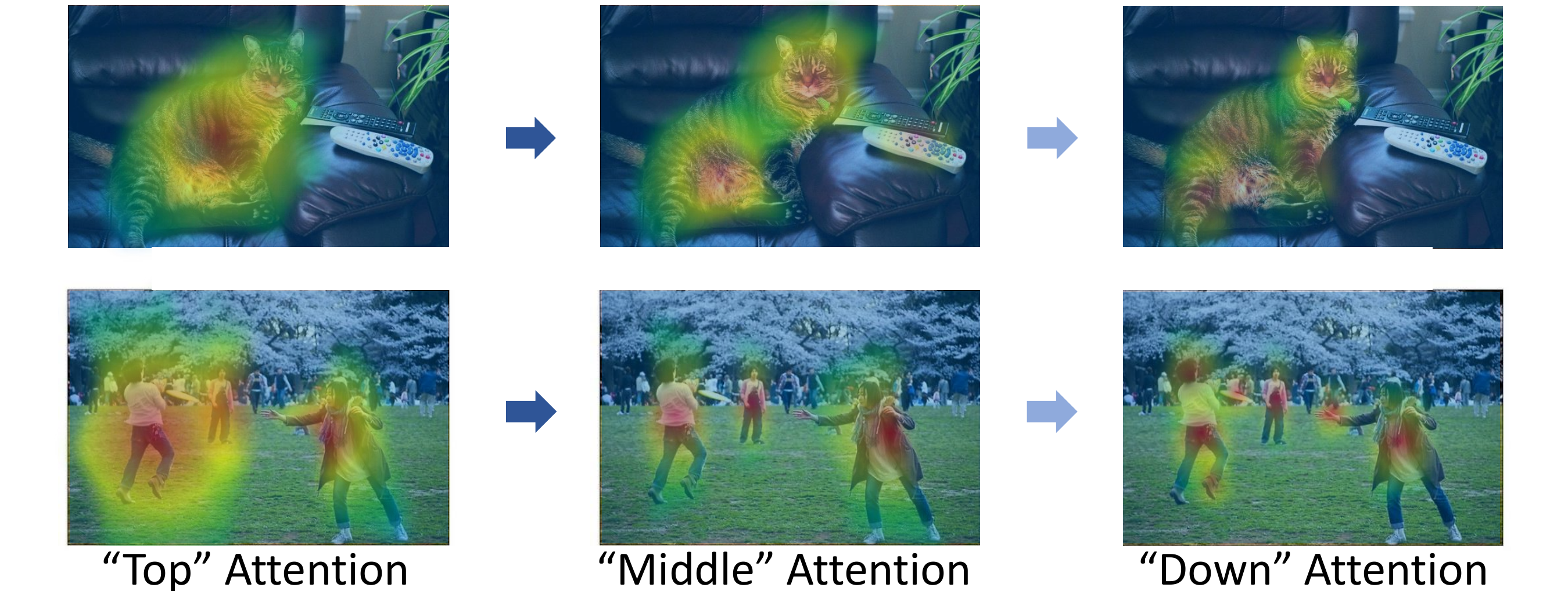}
		\end{center}
		\vspace{-0.11in}
		\caption{
			Top-Down Attentions. The model introduces global information at ``top" attention stage to focus on coarse areas and then gradually adjusts the attention to elaborate areas that contain the most important features. Such a manner can generate fine-grained attention maps with high-level semantics and reduce the loss of information.
		}
		\label{fig:topdownatt}
		\vspace{-0.22in}
	\end{figure}
	
	\begin{figure*}[t]
		\begin{center}
			\includegraphics[width=0.9\linewidth, height=2.8in]{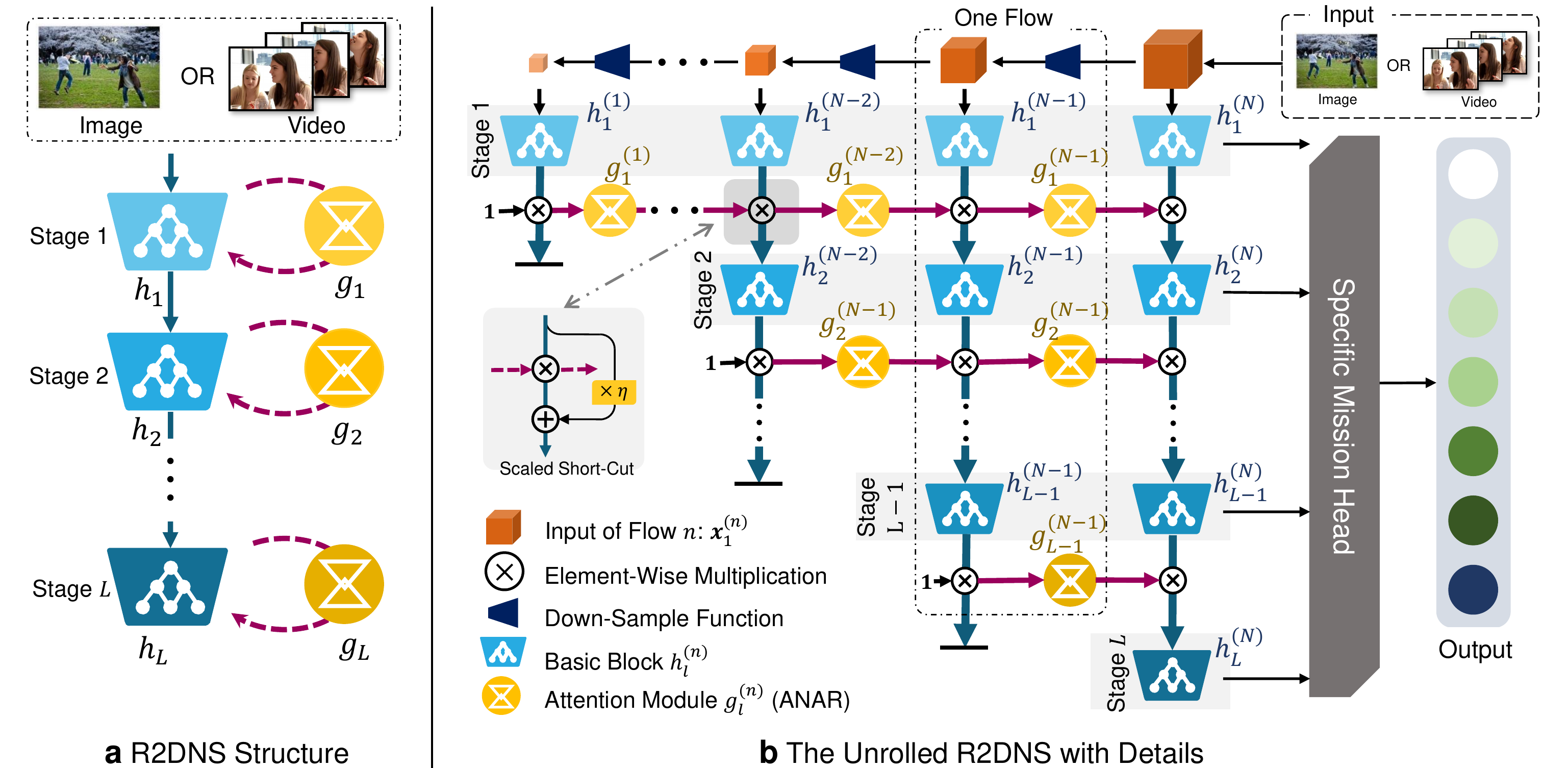}
		\end{center}
		\vspace{-0.1in}
		\caption{Structure of R2DNS. \textbf{a}) The R2DNS consists of multiple stages and at each stage, there is a spatial feature extractor $h_l[\cdot]$ and an attention module $g_l[\cdot]$ (ANAR). The dashed line denotes that it is a path across the timestamps. \textbf{b}) R2DNS first samples the original input down to different scales as  inputs of each recursive stamp (vertical flow). $h_l[\cdot]$ and $g_l[\cdot]$ share the same parameters among different flows. The attention module outputs ``top" attention maps and multiplies them on ``down" features to form the top-down flows in horizontal paths as the red lines show, while in vertical flows, the spatial blocks extract basic features in a bottom-up manner as the blue lines show. Note that the front vertical flows are shorter than the hind ones}
		\label{fig:MSRS}
		\vspace{-0.2in}
	\end{figure*}
	
	In this paper, to extract the top-down attention features (see Fig.~\ref{fig:topdownatt}), we cannot simply extract the attention along with the basic features, instead, we utilize a recursive structure. By adding the recursive dimension, the proposed model can establish the top-down features step by step. It consists of two principal techniques, namely \textit{Recursive Dual-Directional Nested Structure} (R2DNS) and \textit{Attention Network across Recurrence} (ANAR). As its name, R2DNS is a recursive structure and information flows in two nested directions: recursive and structural directions. At each recursive step, it takes as input the images with different scales, from small to large. The small scale inputs contain global but coarse information while the large ones contain local but detailed information. Thus, in the recursive direction, information flows from ``global" to ``local", forming the top-down path. While in the structural direction, the multi-layer deep model still works in the bottom-up scheme. Therefore, R2DNS works in a mixed top-down and bottom-up manner. 
	
	Attention Network across Recurrence (ANAR) serves as a recursive bridge and takes charge of propagating attention features along the recursive paths of R2DNS. It is an hourglass-like network which has been successfully applied to human pose estimation~\cite{newell2016stacked,fang2017rmpe,xiu2018pose} and image segmentation~\cite{long2015fully,fang2019instaboost}. 
	We adopt such a structure to compute attention maps based on features of previous recursive step and then add them as soft weights on next step's spatial features. Thus, it is ANAR that acts as the connection bridges of the recursive steps, instead of the hidden states and it deep nests top-down attention to bottom-up features to work cooperatively.
	
	We conduct comprehensive experiments on several tasks to evaluate our proposed Top-Down Attention Framework. Results reveal that: 1) The ANAR can generate effective attention maps with top-down characteristics. 2) The R2DNS is easy to train in the end-to-end setting. 3) The Top-Down Attention Framework can enjoy accuracy gained from the mixed top-down and bottom-up features, greatly surpassing the corresponding baselines and other attention methods.
	
	We evaluate our framework on several visual tasks: image classification on CIFAR-10~\cite{krizhevsky2009learning} and ImageNet~\cite{russakovsky2015imagenet}, action recognition on Kinetics~\cite{kay2017kinetics}, objection detection, and human pose estimation on COCO~\cite{lin2014microsoft}. For image classification and action recognition, TDAF modified on ResNet50~\cite{he2016deep} and 3D-ResNet50~\cite{hara3dcnns} achieves performances better than ResNet152 and 3D-ResNet101. For object detection, TDAF improves FCOS's~\cite{tian2019fcos} performance by 2.7\% AP. For pose estimation, the performances improve 1.6\% AP on the SimplePose~\cite{xiao2018simple}.

	\vspace{-0.1in}
	\section{Related Work}
	\vspace{-0.03in}
	\noindent\textbf{Attention Mechanisms}
	Attention mechanisms widely exist in the information processing system of human brains~\cite{buschman2007top,corbetta2002control}. Many recent works have incorporated them into artificial neural networks. In sequential problems of NLP~\cite{bahdanau2014neural,vaswani2017attention,lin2017structured,xu2015show}, attention mechanisms are widely adopted in recurrent neural networks (RNN)~\cite{pang2019deep}, Long Short Term Memory (LSTM)~\cite{Hochreiter1997LongSM}, SCS~\cite{pang2020complex}, and Transformer~\cite{vaswani2017attention} to capture the relationships between words or sentences. In computer vision, many tasks like fine-grained recognition~\cite{fu2017look,wang2015multiple,fang2018pairwise,pang2020adverb}, image captioning~\cite{anderson2018bottom,anne2016deep,xu2015show}, classification~\cite{mnih2014recurrent,hu2018squeeze,woo2018cbam,wang2017residual,tang2020asynchronous}, and segmentation~\cite{ren2017end,chen2016attention,cao2020asap} also utilize attention mechanisms based on soft attention maps or bounding-boxes to search salient areas. Moreover, self-attention structures~\cite{wang2018non,zhu2019empirical,huang2018ccnet,dai2019transformer} focusing on the combination weight of elements (pixels in vision) are another attention method that adopts adjacent matrix to present attentions.
	
	The above attention mechanisms work in the bottom-up pattern. They capture attentions first from local information, leading to neglect of context information~\cite{Oliva2003et}. To this end, we propose the new recurrent top-down attention which is closer to human's attention~\cite{corbetta2002control,buschman2007top}, and inchoate attempts~\cite{Oliva2003et} have proved its effectiveness.
	
	\noindent\textbf{Multi-Scale Structures}
	Multi-scale structures which have great scale invariance are widely adopted in vision models. In~\cite{chen2016attention,Piao_2019_ICCV,tian2019fcos,Lin2017Focal,pang2020tubetk}, different scales take charge of detecting the object with different sizes. DLA~\cite{YuDeep} aggregates features from different layers to form more efficient ones. SlowFast network~\cite{feichtenhofer2018slowfast} adopts different scales both in the spatial dimension and temporal dimension to split the works of temporal and spatial feature extracting and reduce the resource consumption. HRNet~\cite{sun2019deep} consists of several parallel flows with inputs of different scales, which aims at maintaining the high-resolution feature map across the whole network, instead of regaining them from the low-resolution map.
	
	In our model, multi-scale inputs are not treated equally like previous works. The attention information flows only from small scale inputs to large ones. Because the attention features from small scale inputs are used as soft masks added onto large ones, these attention features can also be treated as control gates working on the spatial features. 
	
	\vspace{-0.1in}
	\section{Approach}
	\vspace{-0.03in}
	To imitate human brains that observe environment with a top-down attention mechanism, we propose the Top-Down Attention Framework (TDAF) with a recursive structure. In this section, we will introduce its two main parts in turn: 1) the Recursive Dual-Directional Nested Structure (R2DNS) designed to form the mixed top-down and bottom-up information paths in dual directions; 2)~the Attention Network across Recurrence (ANAR) designed to generate and propagate top-down attention along top-down (recursive) paths.
	
	\vspace{-0.07in}
	\subsection{Recursive Dual-Directional Nested Structure}
	\vspace{-0.03in}
	Inspired from the working manner of human attention, we need to build top-down information flow paths in the model, which gradually form attentions from coarse to detailed. 
	
	\noindent\textbf{Challenge} It is not easy to endow the existing deep vision models a top-down information flow, as the stacked CNN layers need to combine low-level local features into high-level global ones and this is a bottom-up process. Adopting residual or parallel modules like~\cite{he2016deep,wang2017residual,zhang2018image} can add more flows but they still behave like traditional CNN, working in a bottom-up manner. Adding a reverse flow from high layers like~\cite{Lin2016Feature,Lin2017Focal,tian2019fcos} cannot combine the bottom-up spatial features and top-down attention maps.
	
	\noindent\textbf{Overall Structure}
	Therefore, we go down to consider how to form the top-down path and keep the high-level spatial features simultaneously.
	For this, we design the Recursive Dual-Directional Nested Structure (R2DNS). It adds another dimension, the recursive dimension, into the traditional bottom-up networks, with which the model can form orthogonal information paths in recursive and structural directions marked as horizontal paths (red lines) and vertical flows (blue lines) in Fig.~\ref{fig:MSRS} \textbf{b}. The same as other CNN networks, vertical flows are bottom-up ones which are responsible for capturing deep spatial representations. And in horizontal direction, the recursive paths extract top-down attention features with a tailored multi-scale mechanism.
	
	As shown in Fig.~\ref{fig:MSRS} \textbf{a}, in vertical flows, the computing units are $\{h_l\}$, responsible for capturing deep spatial representations as other CNN networks, where $l \in \{1,2,...,L\}$ and $L$ is the number of stages in R2DNS. Besides $\{h_l\}$, we design recursive modules $\{g_l\}$. They take spatial features of the last vertical flow as input to generate attention maps and map them onto the next flow. These recursive operations between adjacent flows form the horizontal information paths.
	
	\noindent\textbf{Vertical Flows}
	The unrolled and detailed calculation procedure is shown in Fig.~\ref{fig:MSRS} \textbf{b}. In each vertical flow, the network is a typical deep CNN model with multiple stages: $\{h_l | l \in\{0,1,...,L\}\}$. It can be implemented based on popular image or video backbones. Note that we apply the asymmetric recursive structure: the front vertical flows are shorter (have fewer stages) than the hind ones. With an original backbone having $L$ stages, the number of stages $S$ in vertical flow $n$  can be formalized as:
	\vspace{-0.05in}
	\begin{equation}
	S^{(n)} = L-(N-n)~,
	\vspace{-0.05in}
	\end{equation}
	
	where $N\leq L$ is the total number of flows. And weights are shared among the vertical flows.
	
	In order to form top-down information along the horizontal paths, we adopt a down-sample function (DSF) to generate the input $\mathbf{x}_1$ of each vertical flow $n$ as:
	\vspace{-0.05in}
	\begin{equation}
	\begin{aligned}
	\mathbf{x}_1^{(N)} & = \mathbf{X} \\
	\mathbf{x}_1^{(n)} & = {\rm downsample}[\mathbf{x}_1^{(n+1)}]~,
	\end{aligned}
	\vspace{-0.05in}
	\end{equation}
	where $\mathbf{X}$ is the original input of the whole framework. DSF removes details, therefore, the same receptive field will contain more global information after DSF. This setting makes front vertical flows access to global but coarse information while hind flows get local and detailed one, which form top-down flows in horizontal paths. 
	
	\begin{figure*}[t]
		\begin{center}
			\includegraphics[width=\linewidth, height=1.8in]{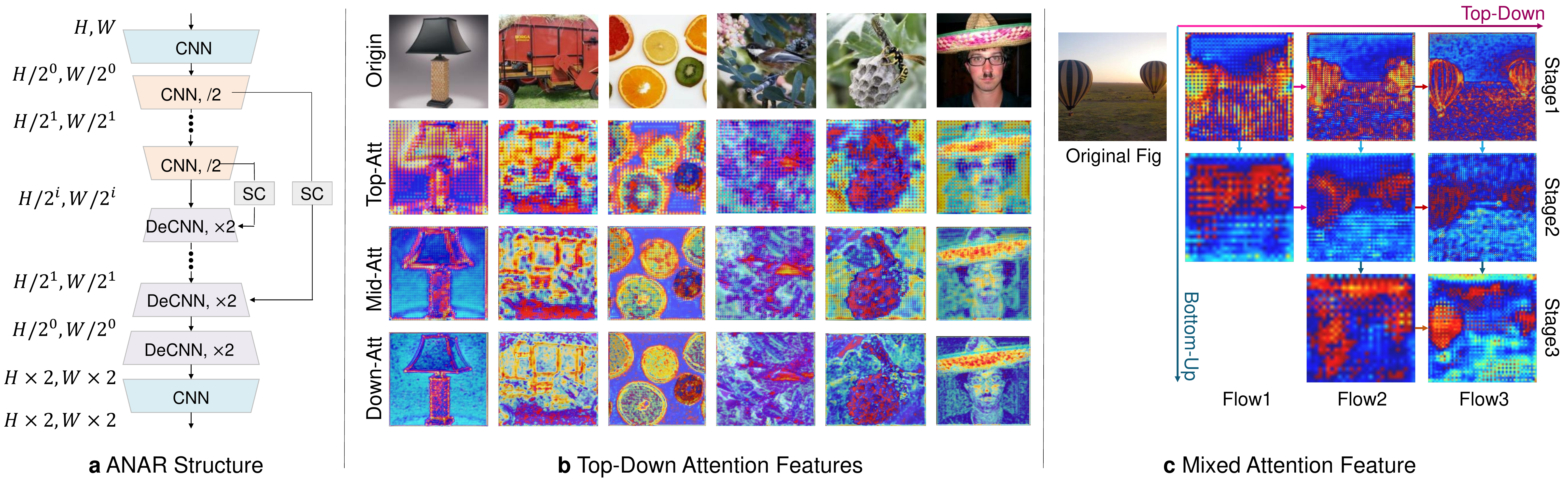}
		\end{center}
		\vspace{-0.1in}
		\caption{\textbf{a}) Structure of ANAR. As an hourglass module, ANAR first reduces the resolution to enlarge the receptive field. Then DeCNN layers are adopted to map the features to the same size of basic features in the next flow. ``SC", a single layer CNN, denotes the skip connects which assist to capture the multi-scale information. \textbf{b}) Attention feature maps with different degrees of fineness generated in different recursive flows. As expected, the front flows capture coarse attention heat-maps to form the ``top" attention and the following flows refine them to more accurate ones and gradually form the ``down" attention. This can reduce the information loss caused by the inaccurate attention generated by a single-step. Here we only exhibit the maps of the first stage. \textbf{c}) Besides the top-down attention in the recursive direction, the deep model also forms the bottom-up feature maps along the structural direction}
		\label{fig:TDAN}
		\vspace{-0.2in}
	\end{figure*}
	
	\noindent\textbf{Horizontal Paths}
	In the horizontal direction, $g_l$ takes global coarse spatial features from the last vertical flow as input, generates attention maps based on them, and maps the attention onto local detailed spatial features of next flow. These recursive operations link up the vertical flows to form new horizontal information paths, and at each stage of R2DNS, due to the global-to-local setting, a series of attention maps from ``top" to ``down" are formed in these paths. Note that weights of $g_l$ are shared among the recursive steps. The detailed structure of $g_l$ is introduced in Sec.~\ref{sec:anar}.
	
	\noindent\textbf{Formal Representation}
	The calculation procedure can be written as:
	\vspace{-0.05in}
	\begin{equation}\label{eq:recursive}
	\begin{aligned}
	&\mathbf{x}_{l+1}^{(n)} = h_{l}[\mathbf{x}_{l}^{(n)}] * g_l[\mathbf{x}_{l+1}^{(n-1)}]~,
	\end{aligned}
	\vspace{-0.05in}
	\end{equation}
	where $\mathbf{x}_{l}^{(n)}$ is the input of the $l$th stage in the $n$th flow and $*$ is the element-wise multiplication. Note that at the start of each horizontal path, the attention feature $g_l[\mathbf{x}_{l+1}^{(n-1)}]$ is initialized to $\mathbf{1}$ as Fig.~\ref{fig:MSRS} \textbf{b} shows. From Eq.~\ref{eq:recursive}, we can see that $\mathbf{x}_{l+1}^{(n)}$ is generated by merging the output of the spatial block $h_l$ in the last stage and the output of attention module $g_l$ in the front flow. As $h_l[\cdot]$ extracts bottom-up features, and $g_l[\cdot]$ works in a top-down manner, the R2DNS is a mixed top-down and bottom-up model. $g_l[\cdot]$ outputs attention masks and they are multiplied onto the basic CNN features generated by $h_l[\cdot]$. These two different information flows transfer in different directions and they are nested deeply at each element-wise multiplication junction.
	
	Let's take a fresh look at the asymmetric recursive structure. From the viewpoint of vertical flows, inputs of front flow with small scales and fewer details do not need deep models. While for horizontal paths, high-level spatial features that are more abstract and have smaller scales do not need long attention chains to form top-down attention features gradually. Thus, this asymmetric structure can reduce unnecessary calculations.
	
	\noindent\textbf{Multi-Flow Batch Normalization}
	Batch normalization (BN)~\cite{Ioffe2015Batch} is a powerful module designed to facilitate the training process of deep model.
	As a recursive structure, multiple vertical flows share the same batch normalization module at each stage. However, the activation of different flows does not share the same distribution. Because the hind flows are masked by the front ones' attention maps, they usually have smaller mean. To this end, we propose the Multi-Flow Batch Normalization (MFBN), which shares the affine parameters among the vertical flows, while for each flow, it has independent mean and variance:
	\vspace{-0.05in}
	\begin{equation}
	\begin{aligned}
	\hat{x}^{(n)}_i &= \frac{x^{(n)}_i - \mu_{\beta^{(n)}}}{\sqrt{\sigma_{\beta^{(n)}}^2 + \epsilon}} \\
	y^{(n)}_i &= \gamma \hat{x}^{(n)}_i + \alpha \equiv {\rm MFBN}_{\gamma,\alpha}(x_i^{(n)})~,
	\end{aligned}
	\vspace{-0.05in}
	\end{equation}
	where $\beta^{(n)} = {x_{1...m}^{(n)}}$ is a mini-batch of $x^{(n)}$, $\gamma$ and $\alpha$ are the affine parameters shared among the flows. The mean $\mu_{\beta^{(n)}}$ and variance $\sigma_{\beta^{(n)}}$ are independent among the flows. 
	
	\vspace{-0.07in}
	\subsection{Attention Network across Recurrence} \label{sec:anar}
	\vspace{-0.03in}
	Attention Network across Recurrence (ANAR) is the computing module responsible for transferring information among the recursive flows of R2DNS. It is also the attention module ($g_l$) that generates soft attention maps and masks them on basic spatial features outputted by $h_l[\cdot]$. This soft attention mechanism can be seen as a control gate of spatial features like~\cite{srivastava2015highway,wang2017residual,pang2019deep}, aiming at strengthening foreground features and weakening background ones.
	
	\noindent\textbf{Structure}
	Extracting attention feature maps is a pixel-level task, thus we design ANAR following the hourglass network~\cite{newell2016stacked} which is usually used in pose estimation. The detailed structure is shown in Fig.~\ref{fig:TDAN} \textbf{a}. In the first half, we apply CNN with stride 2 to reduce the feature space and extract the attention related information. Transposed CNN (DeCNN) is adopted in the second half to map attention maps back to the same size of basic features in the next flow. Like~\cite{newell2016stacked,wang2017residual}, we also add skip connections between the corresponding down-sample and up-sample parts to maintain the multi-scale spatial information. The final activation of ANAR is the sigmoid function which converts feature maps to soft attention masks. To fit different backbones and balance performances with computational requirements, we propose three ANAR structures with different depths. Details are shown in Tab.~\ref{tab:tdan}.
	
	Due to the sigmoid activation function which generates values less than 1, the soft attention mask can only reduce the strength of basic spatial activation. In a deep nested and recursive structure, this property will largely degrade the value of final features. To deal with these two problems, we add a scaled shortcut connection based on Eq.~\ref{eq:recursive} as the following equation shows.
	\vspace{-0.05in}
	\begin{equation}\label{eq:residual}
	\small
	\begin{aligned}
	\mathbf{x}_{l+1}^{(n)} &= h_{l}[\mathbf{x}_{l}^{(n)}] * g_l[\mathbf{x}_{l+1}^{(n-1)}] + \eta * h_{l}[\mathbf{x}_{l}^{(n)}]\\
	&=h_{l}[\mathbf{x}_{l}^{(n)}] * (g_l[\mathbf{x}_{l+1}^{(n-1)}] + \eta)~,
	\end{aligned}
	\vspace{-0.05in}
	\end{equation}
	where $\eta\in[0,1]$ is the shortcut parameter. When $\eta$ is set as 1, this shortcut connection forms a residual learning structure~\cite{he2016deep}, which will solve the problem of activation value drop. However, if $\eta=1$, $(g_l[\mathbf{x}_{l+1}^{(n-1)}] + \eta)\in (1,2)$ holds true, therefore, the soft mask instead can only enhance the spatial activation. In order to both enhance useful features and suppress noise features, we set $\eta$ as 0.5, and we call this ``half-scale" residual learning.
	
	In summary, ANAR forms a series of top-down attention features by a group of recursive operations: firstly generate small coarse attention maps based on the global information, then map this attention onto the more detailed basic spatial feature in the next flow and again, utilize this mapped spatial feature to generate larger and more detailed attention maps. Different from existing attention works~\cite{wang2017residual,xu2015show}, besides the multi-level attention formed alone the structural direction, our model also outputs multiple attention maps with different degrees of fineness (bottom-up and top-down attentions in Fig.~\ref{fig:TDAN} \textbf{b}, \textbf{c}). This process can significantly reduce the information loss caused by the one-step generated inaccurate attention.
	
	\begin{table}[t]
		\setlength{\tabcolsep}{1mm}
		\caption{Detailed settings of ANAR architecture. The structure is shown in Fig.~\ref{fig:TDAN} \textbf{a}. $c$ denotes the number of input channels which depends on $h_l[\cdot]$, ``Down-SP" denotes down-sample and ``/2" means the stride is 2. We adopt MFBN after convolutional layers. Except for the last layer which adopts sigmoid function, ReLU is the activation function of each CNN and DeCNN layer}
		\vspace{-0.15in}
		\renewcommand{\arraystretch}{0.9}
		\newcommand{\tabincell}[2]{\begin{tabular}{@{}#1@{}}#2\end{tabular}}
		\begin{center}
			\scriptsize
			\begin{tabular}{c|c|c|c}
				\hline
				Stage & ANAR-7 & ANAR-5 & ANAR-3 \\
				\hline
				\hline
				Trans (CNN) & 1$\times$1, $c/4$ & 1$\times$1, $c/4$ & 1$\times$1, $c/8$\\ 
				\hline
				Down-SP (CNN) & \tabincell{c}{3$\times$3, $c/8$, /2\\ 3$\times$3, $c/8$, /2} & 3$\times$3, $c/8$, /2  & - \\
				\hline
				Up-SP (DeCNN) & \tabincell{c}{4$\times$4, $c/16$, /2 \\ 4$\times$4, $c/16$, /2 \\ 4$\times$4, $c/32$, /2} & \tabincell{c}{4$\times$4, $c/16$, /2 \\ 4$\times$4, $c/32$, /2} & 4$\times$4, $c/32$, /2 \\
				\hline
				Trans (CNN) & 1$\times$1, $1$ & 1$\times$1, $1$ & 1$\times$1, $1$\\
				\hline
			\end{tabular}
		\end{center}
		\label{tab:tdan}
		\vspace{-0.25in}
	\end{table}
	
	\vspace{-0.07in}
	\subsection{Comparsion with Other Attention Models}\label{sec:SENet}
	\vspace{-0.03in}
	
	Main stream attention mechanisms can be divided into soft attention~\cite{hu2018squeeze,woo2018cbam,chen2016attention,wang2017residual,Fukui2019Attention} and self-attention~\cite{wang2018non,huang2018ccnet,ZhouTemporal,chen20182,KaiyuCompact}. Of course, there are some attention achieved by detecting bounding-box~\cite{fu2017look,anderson2018bottom,mnih2014recurrent}, but this is not the focus of this paper. The main purpose of conventional bottom-up soft attention is to build a group of new weights that is relevant to the input, instead of the fixed convolutional kernel. One problem of this mechanism is that in low layers the receptive field of attention module is too small to form effective attentions. For self-attention, it replaces convolutional operation by attention mechanism in order to enlarge the receptive field, where the essence is to construct high-level features by calculating the ``relevance" between pixels. But its computing complexity is too large to applied on low layers. While our TDAF aims at forming top-down attention chains, building attention maps from coarse to fine, solving the above problems. In the term of design purpose, the TDAF is a new attention mechanism.
	
	\vspace{-0.1in}
	\section{Experiments}\label{sec:experiment}
	\vspace{-0.03in}
	In this section, we evaluate our TDAF on four vision tasks: image classification, object detection, pose estimation, and action recognition. The results reveal TDAF's great performance and universality.
	
	\begin{table}[t]
		\caption{Performance (Acc) on CIFAR-10 test set. The 2-layer ANAR replaces the up-sample layer in ANAR-3 with a parameter-free interpolation. All the models with attention recur three times, i.e. 3 recursive flows}
		\renewcommand{\arraystretch}{0.9}
		\vspace{-0.15in}
		\begin{center}
			\scriptsize
			\begin{tabular}{c|c|c|c}
				\hline
				Backbone & Attention & Params & Acc \\
				\hline
				\hline
				\multirow{3}{*}{VGG16} & No Att & 138.4M & 92.64\\
				& ANAR-2 & 138.4M & 93.02\\
				& ANAR-3 & 138.5M & 93.43\\
				\hline
				\multirow{2}{*}{VGG19} & No Att & 143.7M & 93.13 \\
				& ANAR-3 & 143.8M & \textbf{93.78}\\
				\hline
				\multirow{2}{*}{Res50} & No Att & 25.6M & 93.62\\
				& ANAR-3 & 29.9M & 94.13\\
				\hline
				\multirow{2}{*}{Res101} & No Att & 44.5M & 93.75 \\
				& ANAR-3 & 48.9M & \textbf{94.21}\\
				\hline
			\end{tabular}
		\end{center}
		\label{tab:cifar_res}
		\vspace{-0.25in}
	\end{table}
	
	\vspace{-0.07in}
	\subsection{Image Classification}
	\vspace{-0.03in}
	We first evaluate TDAF on image classification with CIFAR-10~\cite{krizhevsky2014cifar} and ImageNet-2012~\cite{russakovsky2015imagenet} datasets and compare it with the following baselines: VGG~\cite{simonyan2014very}, ResNet~\cite{he2016deep}, ResNeXt~\cite{xie2017aggregated}, MobileNet~\cite{howard2017mobilenets}, and SENet~\cite{hu2018squeeze}.
	
	\noindent\textbf{Implementation}
	For VGG, we insert 5 ANAR modules after each max-pooling layer to form the recursive attention structure. While for the ResNe(X)t, we insert ANARs after each stage respectively. All the BN layers are replaced by our MFBN. Baseline network and its TDAF counterpart are trained with identical optimization scheme following~\cite{he2016deep}. The parameters are initialized by ``Kaiming initialization" proposed in~\cite{he2015delving}. We use SGD optimizer with 256 mini-batch on 8 GPUs to train.
	
	\begin{table*}[t]
		\caption{\textit{a}) Performance on ImageNet validation set (10-crop test). ``\# L" and ``\# F" denote the number of ANAR's layers and flows (recursive steps). \textit{b}) Comparison of performance improvement (PI) on Top-1 accuracy (1-crop test) and computational overheads on ImageNet between different attention mechanisms}
		\renewcommand{\arraystretch}{0.9}
		\vspace{-0.1in}
		\begin{subtable}[t]{0.55\linewidth}
			\centering
			\scriptsize
			\caption{}
			\begin{tabular}{c|c|c|cc|cc}
				\hline
				\multirow{2}{*}{Backbone} & \multicolumn{2}{c|}{ANAR} & \multirow{2}{*}{Params} & \multirow{2}{*}{GFlops} & \multicolumn{2}{c}{Accuracy} \\
				\cline{2-3}
				\cline{6-7}
				& \# L & \# F & &&Top-1&Top-5\\
				\hline
				\hline
				\multirow{3}{*}{VGG16~\cite{simonyan2014very}} & - & - & 138.4M & 15.7 & 74.7 & 92.0\\
				& 5 & 3 & 138.9M & 21.6 & \textbf{75.9} & \textbf{92.7}\\
				& 5 & 2 & 138.9M & 20.6 & 75.5& 92.4 \\
				\hline
				\multirow{5}{*}{Res50~\cite{he2016deep}}& - & - & 25.6M & 4.1 & 77.1 & 93.3 \\
				& 3 & 3 & 29.9M & 5.4 & 78.4 & 94.1\\
				& 5 & 3 & 30.3M & 5.8 & 78.6 & 94.3\\
				& 7 & 3 & 30.5M & 5.8 & \textbf{78.8} & \textbf{94.5} \\
				& 3 & 4 & 29.9M & 5.4 & 78.7 & 94.4\\
				\hline
				\multirow{3}{*}{Res101~\cite{he2016deep}} & - & - & 44.5M & 7.9 & 78.2 & 93.9\\
				& 3 & 3 & 48.9M & 10.0 & 79.9 & 94.8 \\
				& 3 & 4 & 48.9M & 10.0 & \textbf{80.2} & \textbf{95.0} \\
				\hline
				\multirow{2}{*}{Res152~\cite{he2016deep}} & - & - & 60.2M & 11.6 & 78.6 & 94.3\\
				& 3 & 3 & 64.5M & 14.8 & 80.4 & 95.1\\
				\hline
				\multirow{2}{*}{ResX50~\cite{xie2017aggregated}} & - & - & 25.0M & 4.3 & 79.4 & 94.7\\
				& 3 & 3 & 29.4M & 5.6 & 80.3 & 95.0\\
				\hline
				\multirow{2}{*}{ResX101~\cite{xie2017aggregated}} & - & - & 88.8M & 16.5 & 81.0 & 95.5\\
				& 3 & 3 & 93.1M & 20.8 & 81.8 & 95.8\\
				\hline
				\multirow{3}{*}{MobNets~\cite{howard2017mobilenets}} & - & - & 4.2M & 0.57 & 68.6 & 88.49\\
				& 3 & 2 & 5.1M &  0.71 & 71.0 & 90.01 \\
				&&&&&&\\
				\hline
			\end{tabular} 
			\label{tab:imagenet_res}  
		\end{subtable}
		\begin{subtable}[t]{0.45\linewidth}
			\centering
			\caption{}
			\scriptsize
			\begin{tabular}{c|c|c|c}
				\hline
				Baseline & Model & GFLOPs & PI\\
				\hline
				\hline
				\multirow{2}{*}{VGG16} & TDAF & +37.5\% & +1.3\\
				& ABN~\cite{Fukui2019Attention} & - & +0.7\\
				\hline
				\multirow{5}{*}{Res50} & TDAF & +31.7\% & +1.7\\
				& $\rm A^2$-Net~\cite{chen20182} & +58.5\% & +1.7\\
				& NL~\cite{wang2018non} & +12.5\% & +0.8\\
				
				& ResAtt~\cite{wang2017residual} & +53.6\% & +1.1\\
				& GCNL~\cite{KaiyuCompact} & - & +1.2 \\
				& SAST~\cite{parmar2019stand} & +34.5\% & +1.2\\
				& AAAtt~\cite{bello2019attention} & +1.3\% (29\%) & +1.3\\
				\hline
				\multirow{3}{*}{Res152} & TDAF & +27.5\% & +2.0\\
				& ABN~\cite{Fukui2019Attention} & - & +0.8\\
				& GCNL~\cite{KaiyuCompact} & - & +1.1\\
				\hline
				\multirow{3}{*}{ResX101} & TDAF & +24.5\% & +0.9\\
				& NL~\cite{wang2018non} & +9.9\% & +0.4\\
				& CBAM~\cite{woo2018cbam} & +0.1\% & +0.4\\
				\hline
				\multirow{3}{*}{SENet} & TDAF & +26.0\% & +1.3\\
				& ABN~\cite{Fukui2019Attention} & - & +0.8\\
				& CBAM~\cite{woo2018cbam} & +0.1\% & +0.5\\
				\hline
				\multirow{3}{*}{CBAM} & TDAF & +25.9\% & +0.7\\
				& ABN~\cite{Fukui2019Attention} & - & +0.3\\
				& NL~\cite{wang2018non} & +12.5\% & +0.3\\
				\hline
			\end{tabular}
			\label{tab:compare_w_att}  
		\end{subtable}
		\vspace{-0.15in}
	\end{table*}  
	
	\noindent\textbf{CIFAR-10 and Analysis}
	We first evaluate TDAF on the small dataset CIFAR-10 to reveal its effectiveness. As the resolution is only 32 $\times$ 32, we adopt the 3-layer ANAR. We also replace the DeCNN layer in ANAR-3 with a parameter-free interpolation function to build a much smaller ANAR-2 specifically for this small-scale task. 
	
	The experiment results are shown in Tab.~\ref{tab:cifar_res}. We observe that TDAF consistently improves the performance across different backbones and model depths with a small increase in parameters. VGG19 with attention exceeds baseline by 0.65\%, and ResNet-50 by 0.51\%, even better than the much deeper baseline ResNet-101 with much fewer parameters (29.9M vs. 45.5M). In order to test whether we can get great attention with nearly no parameters in ANAR, we replace ANAR-3's DeCNN layer with a parameter-free interpolation to build a much smaller ANAR-2. Unfortunately, this trial leads to a great drop in performance, revealing the importance of the learnable parameters in ANAR.
	
	\noindent\textbf{ImageNet and Analysis}
	We further analyze different attention settings on ImageNet-2012~\cite{russakovsky2015imagenet}. Training images are resized randomly to [256, 480] with its shorter side and a 224 $\times$ 224 crop is sampled from it or its horizontal flip. With larger input size, we apply deeper ANAR and more recursive steps to maximize its potential.
	
	The results are shown in Tab.~\ref{tab:imagenet_res}. The networks with TDAF outperform all the baselines significantly. Note that even applying the smallest attention setting (ANAR-3 with 3 recursive flows), our model adopting ResNet50 as the backbone exceeds deeper ResNet-101 (78.4\% vs. 78.2\%) with much fewer parameters and FLOPs. And this pattern holds at ResNet101 which outperforms ResNet152 by 1.3\% in Top-1 Accuracy. On ResNeXt50 and ResNeXt101~\cite{xie2017aggregated}, there are still considerable improvements (0.9\% and 0.8\%). For MobileNet~\cite{howard2017mobilenets}, our ANAR adopts depthwise convolutional filters proposed in~\cite{howard2017mobilenets} and the performance is improved by 2.4\%.
	
	As for different attention settings, the results reveal that a deeper ANAR module or more recursive steps will lead to better results due to better abilities to extract attention features and longer top-down attention paths which endows the model larger initial receptive fields.
	
	\noindent\textbf{Performance Comparison of Attention Models}
	Here, we compare our TDAF with other attention models on ImageNet. As shown in table~\ref{tab:compare_w_att}, on different baselines, TDAF is competitive among all the mechanisms, achieving great performance boost with relatively small Flops overheads. As for parameter overhead, TDAF adopts 10\% more parameter on ResNet101 and 7\% on ResNet152. This overhead less than 10\% leads to 2\% Top-1 accuracy improvements. Other attention models like Non-local (20\% overhead, 0.8\% improvements), A$^2$ Net (10\%, 1.7\%), ResAtt (24\%, 1.1\%), ABN (\textgreater20\%, 0.8\%), and GCNL (\textgreater20\%, 1.2\%) adopt more parameters and gains less improvements. For each unit of parameters increase, our model delivers 3$\sim$5 times performance increases of Non-Local, ResAtt, ABN, and GCNL. Note that compared with CBAM which directly generates maps based on the original convolutional features, which cannot be adjusted to top-down attention, TDAF introduces more overhead like other attention methods. To further demonstrate the performance of TDAF, we treat CBAM as an advanced baseline. After adopting TDAF, it achieves a considerable performance boost, revealing that the top-down and bottom-up soft attentions can work cooperatively and propel each other to better performances.
	
	\vspace{-0.07in}
	\subsection{Object Detection}
	\vspace{-0.03in}
	We also evaluate TDAF on detection task with the COCO-2017 dataset~\cite{lin2014microsoft}. We choose the one-stage anchor-free model FCOS~\cite{tian2019fcos} as baseline. 
	
	\noindent\textbf{Implementation}
	Still, the $\rm downsample$ stride is 2. We modify the backbone of FCOS with our TDAF. The data pre-processing and training methods follow the original FCOS (``2$\times$" training schedule with multi-scale input). The weights of all the backbones are initialized by the parameters pre-trained on the ImageNet dataset.
	
	\begin{table}[t]
		\setlength{\tabcolsep}{1mm}
		\caption{Performance and training curve on COCO-2017. ``4F" denotes the number of recursive flows is 4}
		\vspace{-0.15in}
		\renewcommand{\arraystretch}{0.9}
		\begin{center}
			\scriptsize
			\begin{tabular}{l|l|cccc|c}
				\hline
				Backbone & Top-Down Att & AP & $\text{AP}_S$ & $\text{AP}_M$ & $\text{AP}_L$ & $\text{Params}$ \\
				\hline\hline
				\multirow{3}{*}{ResNet50~\cite{he2016deep}} & No Att & 37.1 & 21.3 & 41.0 & 47.8 & 32.244M \\ 
				& ANAR-3, 3F & 39.6 & 23.9 & \textbf{43.4} & 51.1 & 32.462M \\
				& ANAR-5, 3F & \textbf{39.8} & \textbf{24.0} & 43.3 & \textbf{51.2}  & 32.885M \\
				\hline
				\multirow{4}{*}{ResNet101~\cite{he2016deep}} & No Att & 41.5 & 24.4 & 44.8 & 51.6 & 51.184M \\
				& ANAR-3, 3F & 42.0 & \textbf{25.5} & 46.4 & 54.4 & 51.454M \\
				& ANAR-3, 4F & \textbf{42.3} & 25.3 & \textbf{46.7} & 55.1 & 51.454M \\
				& ANAR-5, 3F & 42.2 & 25.3 & 46.6 & \textbf{55.3} &51.877M\\
				\hline
			\end{tabular}
			\includegraphics[width=0.9\linewidth]{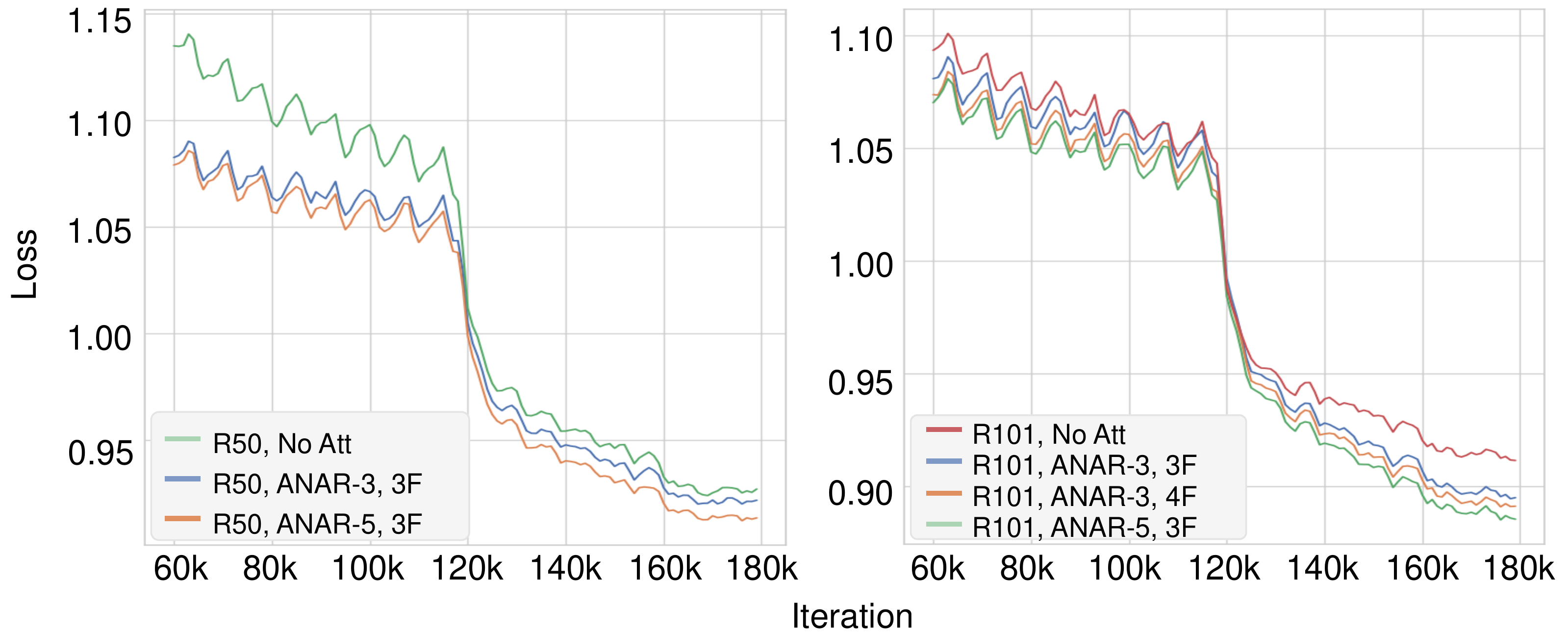}
		\end{center}
		\label{tab:coco_res}
		\vspace{-0.3in}
	\end{table}
	
	Tab.~\ref{tab:coco_res} reports the performances. TDAF-Res50-FCOS outperforms Res50-FOCS by 2.7\% on the standard AP metric (7.3\% relatively). Top-down attention also benefits the deeper Res101-FCOS and achieves 0.9\% improvements.
	
	To provide some insights into the influence of Top-Down Attention Framework on the optimization process, the training curves are depicted in Tab.~\ref{tab:coco_res}. The TDAF yields lower training errors throughout the optimization procedure and more recursive steps or deeper ANAR lead to lower training losses throughout the optimization procedure.
	
	\vspace{-0.07in}
	\subsection{Pose Estimation}
	\vspace{-0.03in}
	We further evaluate TDAF on pose estimation task using the COCO-2017 dataset. The mainstream methods of this task utilize heat-maps to localize key point positions. To some extent, this is a kind of attention-based solution. We want to evaluate whether our top-down attention mechanism can benefit this already attention-driven task. 
	
	\noindent\textbf{Implementation}
	We adopt SimplePose~\cite{xiao2018simple} as our baseline and modify its deconvolutional layers into our TDAF version with 3 recursive flows. The details of training processes maintain the same with the original baseline. We adopt Faster-RCNN~\cite{girshick2015fast} and YOLO-v3~\cite{redmon2018yolov3} detectors.
	
	The results are shown in Tab.~\ref{tab:pose_res}. 
	With ResNet50 as the backbone, our TDAF improves AP by 1.6\%, while for ResNet101, the improvements are 0.8\%. Considering pose estimation is already based on attention, these promotions are considerable, demonstrating the potential of the top-down manner in attention mechanism. Also, a deeper ANAR structure leads to better performances.
	
	\begin{table}[t]
		\caption{Performance (AP in \%) of the SimplePose~\cite{xiao2018simple} on COCO-2017 validation set. ``3F" denotes 3 recursive flows. ``FRCNN" and ``YOLO" denote Faster-RCNN and YOLO-V3 detectors}
		\renewcommand{\arraystretch}{0.9}
		\vspace{-0.15in}
		\begin{center}
			\scriptsize
			\begin{tabular}{l|l|ccc}
				\hline
				\multirow{2}{*}{Backbone} & \multirow{2}{*}{Top-Down Att} & \multicolumn{3}{c}{Detector} \\
				\cline{3-5}
				& & GT & FRCNN & YOLO\\
				\hline\hline
				\multirow{4}{*}{ResNet50} & No Att & 72.7 & 70.1 & 70.4 \\
				& ANAR-3, 3F & 73.8 & 71.3 & 71.4\\
				& ANAR-5, 3F & 74.1 & 71.5  & 71.5\\
				& ANAR-7, 3F & \textbf{74.3} & \textbf{71.6} & \textbf{71.7} \\
				\hline
				\multirow{3}{*}{ResNet101} & No Att & 73.7& 70.9& 71.3\\
				& ANAR-3, 3F & 74.4 & 71.9 & \textbf{72.4}\\
				& ANAR-5, 3F & \textbf{74.5} & \textbf{72.0} & 72.3\\
				\hline
			\end{tabular}
		\end{center}
		\label{tab:pose_res}
		\vspace{-0.25in}
	\end{table}
	
	\vspace{-0.07in}
	\subsection{Action Recognition}
	\vspace{-0.03in}
	Next, we apply the proposed framework onto the video action recognition task.
	We adopt 3D-ResNet~\cite{hara3dcnns} as our baseline and the experiments are conducted on Kinetics-400 dataset~\cite{kay2017kinetics}.
	
	\noindent\textbf{Implementation}
	In these experiments, the $\rm downsample$ function is conducted on 3 dimensions, where we treat the temporal and spatial dimensions equally, just as the 3D-CNN does. In 3D-ResNet, all the 2D kernels are extended to 3D, therefore, we also adopt a 3D version ANAR.
	Here, the ANAR-5 with 3 recursive steps is utilized. Similar to \cite{Carreira2017Quo}, we inflate the 2D weights pre-trained on ImageNet to initialize the model.
	
	We split the video into clips of 16 frames as input and the frames are resized to 112 $\times$ 112. The other training and inference settings all follow \cite{hara3dcnns}.
	
	Results on the Kinetics dataset shown in Tab.~\ref{tab:Kinetics_res} demonstrate consistent performances of TDAF with the above tasks. It outperforms 3D-ResNet50 by 1.8\% without ImageNet pre-training and 1.2\% with pre-training. The same, TDAF with ResNet50 outperforms the ResNet101 baseline (67.5\% vs. 67.2\%), revealing that the TDAF also has great performances in capturing the spatial and temporal attention simultaneously and can be applied to video tasks.
	
	\begin{table}[t]
		
		\caption{Performance on the Kinetics validation set. We adopt ANAR-5 and 3 recursive flows in our TDAF models}
		\renewcommand{\arraystretch}{0.9}
		\vspace{-0.1in}
		\begin{center}
			\scriptsize
			\begin{tabular}{l|c|c}
				\hline
				Model & Params & Acc (\%) \\
				\hline
				\hline
				3D-ResNet50 & 47.019M & 65.7 \\
				3D-ResNet50, Pre-trained & 47.019M & 68.9\\
				3D-ResNet101 & 86.065M & 67.2 \\
				\hline
				TDAF-3D-ResNet50 & 48.417M & 67.5\\
				TDAF-3D-ResNet50, Pre-trained & 48.417M & \textbf{70.1} \\
				\hline
			\end{tabular}
		\end{center}
		\label{tab:Kinetics_res}
		\vspace{-0.2in}
	\end{table}
	
	\vspace{-0.07in}
	\subsection{Analysis}
	\vspace{-0.03in}
	\noindent\textbf{On Multi-Scale and Attention} Until now, there is a question that whether the great improvements are achieved by our top-down attention mechanism or simply by the multi-scale information. Note that we only feed the features from the last flow into the task heads as shown in Fig.~\ref{fig:MSRS}, thus from the view of the task head, TDAF does not access to the multi-scale spatial features, instead, it only utilizes the attention features from other flows. Even so, we conduct controlled trials on ImageNet to show that it is the top-down attention that works in the whole framework. On ResNet-50, we adopt the same R2DNS setting (3 flows) and remove the attention modules among the recursive steps. In this way the multiple flows are standalone without attention connections. Then the features of all the flows are concatenated to conduct the classification in a multi-scale way. Compared with the corresponding TDAF with ANAR-3, the performance of this multi-scale setting drops 1.1\% Top-1 accuracy on ImageNet, achieving 77.5\%, only 0.4\% improvements over the baseline without multi-scale setting (more results in supplementary files). This reveals that multi-scale information can lead to better performances, but the great improvements are actually achieved by our top-down attention mechanism.

	\noindent\textbf{On Time Overhead}
	One may be concerned that the recurrent structure cannot be parallelled and lead to a long time delay. But actually, only the ANAR parts cannot be parallelled, while the multiple backbone flows can. As Fig. 2b shows, in first stage, we could calculate $h_1^{(1)}$ to $h_1^{(N)}$ parallelly, then calculate $g_1^{(1)}$ to $g_1^{(N-1)}$ recursively, so can other stages. Thus, the total time of our model is just the sum of the time of the original backbone and the attention layers. Because the attention layers and the additional backbone flows are much lighter than the original backbone, the latency overhead is less than 25\% for Res50 and 15\% for Res101. On TitanXP with 256 batch size, the forward time of Res50 and its attention version is 0.015s and 0.018s, and 0.031s vs 0.035s for Res101. Compared with GALA~\cite{linsley2018global} (86\%), CBAM (56\%), and AAAtt (29\%) (these results are published in~\cite{bello2019attention}), this latency is in a reasonable range. For detection, because the head and NMS consume much time, the total time overhead on Res50 is less than 10\%.

	\vspace{-0.1in}
	\section{Conclusion}
	\vspace{-0.03in}
	In this paper, we proposed TDAF to mix top-down attention features with bottom-up spatial features. The first part of our framework is the Recursive Dual-Directional Nested Structure (R2DNS) designed to form the orthogonal top-down and bottom-up information paths and the second part is the Attention Network across Recurrence (ANAR) aiming at extracting attention maps to enhance useful features and suppress noises. Compared with baselines and other attention mechanisms, our TDAF achieves remarkable improvements on image classification, object detection, pose estimation, and action recognition. Comprehensive analyses were presented to further validate different settings of TDAF.
	
	\section{Acknowledgement}
	This work is supported in part by the National Key R\&D Program of China, No. 2017YFA0700800, National Natural Science Foundation of China under Grants 61772332 and Shanghai Qi Zhi Institute, SHEITC (018-RGZN-02046).
	{\small
		\bibliography{egbib}

\begin{thebibliography}{67}
\providecommand{\natexlab}[1]{#1}
\providecommand{\url}[1]{\texttt{#1}}
\providecommand{\urlprefix}{URL }
\expandafter\ifx\csname urlstyle\endcsname\relax
  \providecommand{\doi}[1]{doi:\discretionary{}{}{}#1}\else
  \providecommand{\doi}{doi:\discretionary{}{}{}\begingroup
  \urlstyle{rm}\Url}\fi

\bibitem[{Anderson et~al.(2018)Anderson, He, Buehler, Teney, Johnson, Gould,
  and Zhang}]{anderson2018bottom}
Anderson, P.; He, X.; Buehler, C.; Teney, D.; Johnson, M.; Gould, S.; and
  Zhang, L. 2018.
\newblock Bottom-up and top-down attention for image captioning and visual
  question answering.
\newblock In \emph{CVPR}, 6077--6086.

\bibitem[{Anne~Hendricks et~al.(2016)Anne~Hendricks, Venugopalan, Rohrbach,
  Mooney, Saenko, and Darrell}]{anne2016deep}
Anne~Hendricks, L.; Venugopalan, S.; Rohrbach, M.; Mooney, R.; Saenko, K.; and
  Darrell, T. 2016.
\newblock Deep compositional captioning: Describing novel object categories
  without paired training data.
\newblock In \emph{CVPR}, 1--10.

\bibitem[{Bahdanau, Cho, and Bengio(2014)}]{bahdanau2014neural}
Bahdanau, D.; Cho, K.; and Bengio, Y. 2014.
\newblock Neural machine translation by jointly learning to align and
  translate.
\newblock \emph{arXiv preprint arXiv:1409.0473} .

\bibitem[{Bello et~al.(2019)Bello, Zoph, Vaswani, Shlens, and
  Le}]{bello2019attention}
Bello, I.; Zoph, B.; Vaswani, A.; Shlens, J.; and Le, Q.~V. 2019.
\newblock Attention augmented convolutional networks.
\newblock In \emph{ICCV}, 3286--3295.

\bibitem[{Buschman and Miller(2007)}]{buschman2007top}
Buschman, T.~J.; and Miller, E.~K. 2007.
\newblock Top-down versus bottom-up control of attention in the prefrontal and
  posterior parietal cortices.
\newblock \emph{science} 315(5820): 1860--1862.

\bibitem[{Cao et~al.(2020)Cao, Lu, Lu, Pang, Liu, and Yuille}]{cao2020asap}
Cao, H.; Lu, Y.; Lu, C.; Pang, B.; Liu, G.; and Yuille, A. 2020.
\newblock ASAP-Net: Attention and Structure Aware Point Cloud Sequence
  Segmentation.
\newblock \emph{arXiv preprint arXiv:2008.05149} .

\bibitem[{Carreira and Zisserman(2017)}]{Carreira2017Quo}
Carreira, J.; and Zisserman, A. 2017.
\newblock Quo Vadis, Action Recognition? A New Model and the Kinetics Dataset .

\bibitem[{Chen et~al.(2016)Chen, Yang, Wang, Xu, and
  Yuille}]{chen2016attention}
Chen, L.-C.; Yang, Y.; Wang, J.; Xu, W.; and Yuille, A.~L. 2016.
\newblock Attention to scale: Scale-aware semantic image segmentation.
\newblock In \emph{CVPR}, 3640--3649.

\bibitem[{Chen et~al.(2018)Chen, Kalantidis, Li, Yan, and Feng}]{chen20182}
Chen, Y.; Kalantidis, Y.; Li, J.; Yan, S.; and Feng, J. 2018.
\newblock A\^{} 2-nets: Double attention networks.
\newblock In \emph{NeurIPS}, 352--361.

\bibitem[{Corbetta and Shulman(2002)}]{corbetta2002control}
Corbetta, M.; and Shulman, G.~L. 2002.
\newblock Control of goal-directed and stimulus-driven attention in the brain.
\newblock \emph{Nature reviews neuroscience} 3(3): 201.

\bibitem[{Dai et~al.(2019)Dai, Yang, Yang, Cohen, Carbonell, Le, and
  Salakhutdinov}]{dai2019transformer}
Dai, Z.; Yang, Z.; Yang, Y.; Cohen, W.~W.; Carbonell, J.; Le, Q.~V.; and
  Salakhutdinov, R. 2019.
\newblock Transformer-xl: Attentive language models beyond a fixed-length
  context.
\newblock \emph{arXiv preprint arXiv:1901.02860} .

\bibitem[{Fang et~al.(2018)Fang, Cao, Tai, and Lu}]{fang2018pairwise}
Fang, H.-S.; Cao, J.; Tai, Y.-W.; and Lu, C. 2018.
\newblock Pairwise body-part attention for recognizing human-object
  interactions.
\newblock In \emph{ECCV}, 51--67.

\bibitem[{Fang et~al.(2019)Fang, Sun, Wang, Gou, Li, and
  Lu}]{fang2019instaboost}
Fang, H.-S.; Sun, J.; Wang, R.; Gou, M.; Li, Y.-L.; and Lu, C. 2019.
\newblock Instaboost: Boosting instance segmentation via probability map guided
  copy-pasting.
\newblock In \emph{Proceedings of the IEEE International Conference on Computer
  Vision}, 682--691.

\bibitem[{Fang et~al.(2017)Fang, Xie, Tai, and Lu}]{fang2017rmpe}
Fang, H.-S.; Xie, S.; Tai, Y.-W.; and Lu, C. 2017.
\newblock Rmpe: Regional multi-person pose estimation.
\newblock In \emph{ICCV}, 2334--2343.

\bibitem[{Feichtenhofer et~al.(2018)Feichtenhofer, Fan, Malik, and
  He}]{feichtenhofer2018slowfast}
Feichtenhofer, C.; Fan, H.; Malik, J.; and He, K. 2018.
\newblock Slowfast networks for video recognition.
\newblock \emph{arXiv preprint arXiv:1812.03982} .

\bibitem[{Fu, Zheng, and Mei(2017)}]{fu2017look}
Fu, J.; Zheng, H.; and Mei, T. 2017.
\newblock Look closer to see better: Recurrent attention convolutional neural
  network for fine-grained image recognition.
\newblock In \emph{CVPR}, 4438--4446.

\bibitem[{Fukui and et~al.(2019)}]{Fukui2019Attention}
Fukui, H.; and et~al. 2019.
\newblock Attention Branch Network: Learning of Attention Mechanism for Visual
  Explanation arXiv:1812.10025v2.
\newblock In \emph{CVPR}.

\bibitem[{Girshick(2015)}]{girshick2015fast}
Girshick, R. 2015.
\newblock Fast r-cnn.
\newblock In \emph{ICCV}, 1440--1448.

\bibitem[{Hara, Kataoka, and Satoh(2018)}]{hara3dcnns}
Hara, K.; Kataoka, H.; and Satoh, Y. 2018.
\newblock Can Spatiotemporal 3D CNNs Retrace the History of 2D CNNs and
  ImageNet?
\newblock In \emph{CVPR}, 6546--6555.

\bibitem[{He et~al.(2015)He, Zhang, Ren, and Sun}]{he2015delving}
He, K.; Zhang, X.; Ren, S.; and Sun, J. 2015.
\newblock Delving deep into rectifiers: Surpassing human-level performance on
  imagenet classification.
\newblock In \emph{ICCV}, 1026--1034.

\bibitem[{He et~al.(2016)He, Zhang, Ren, and Sun}]{he2016deep}
He, K.; Zhang, X.; Ren, S.; and Sun, J. 2016.
\newblock Deep residual learning for image recognition.
\newblock In \emph{CVPR}, 770--778.

\bibitem[{Hochreiter and Schmidhuber(1997)}]{Hochreiter1997LongSM}
Hochreiter, S.; and Schmidhuber, J. 1997.
\newblock Long Short-Term Memory.
\newblock \emph{Neural Computation} 9: 1735--1780.

\bibitem[{Howard et~al.(2017)Howard, Zhu, Chen, Kalenichenko, Wang, Weyand,
  Andreetto, and Adam}]{howard2017mobilenets}
Howard, A.~G.; Zhu, M.; Chen, B.; Kalenichenko, D.; Wang, W.; Weyand, T.;
  Andreetto, M.; and Adam, H. 2017.
\newblock Mobilenets: Efficient convolutional neural networks for mobile vision
  applications.
\newblock \emph{arXiv preprint} .

\bibitem[{Hu, Shen, and Sun(2018)}]{hu2018squeeze}
Hu, J.; Shen, L.; and Sun, G. 2018.
\newblock Squeeze-and-excitation networks.
\newblock In \emph{CVPR}, 7132--7141.

\bibitem[{Huang et~al.(2018)Huang, Wang, Huang, Huang, Wei, and
  Liu}]{huang2018ccnet}
Huang, Z.; Wang, X.; Huang, L.; Huang, C.; Wei, Y.; and Liu, W. 2018.
\newblock Ccnet: Criss-cross attention for semantic segmentation.
\newblock \emph{arXiv preprint arXiv:1811.11721} .

\bibitem[{Ioffe and Szegedy(2015)}]{Ioffe2015Batch}
Ioffe, S.; and Szegedy, C. 2015.
\newblock Batch Normalization: Accelerating Deep Network Training by Reducing
  Internal Covariate Shift .

\bibitem[{Kay et~al.(2017)Kay, Carreira, Simonyan, Zhang, Hillier,
  Vijayanarasimhan, Viola, Green, Back, Natsev et~al.}]{kay2017kinetics}
Kay, W.; Carreira, J.; Simonyan, K.; Zhang, B.; Hillier, C.; Vijayanarasimhan,
  S.; Viola, F.; Green, T.; Back, T.; Natsev, P.; et~al. 2017.
\newblock The kinetics human action video dataset.
\newblock \emph{arXiv preprint} .

\bibitem[{Krizhevsky, Hinton et~al.(2009)}]{krizhevsky2009learning}
Krizhevsky, A.; Hinton, G.; et~al. 2009.
\newblock Learning multiple layers of features from tiny images.
\newblock Technical report, Citeseer.

\bibitem[{Krizhevsky, Nair, and Hinton(2014)}]{krizhevsky2014cifar}
Krizhevsky, A.; Nair, V.; and Hinton, G. 2014.
\newblock The CIFAR-10 dataset.
\newblock \emph{online: http://www. cs. toronto. edu/kriz/cifar. html} 55.

\bibitem[{Li, Zhu, and Gong(2018)}]{li2018harmonious}
Li, W.; Zhu, X.; and Gong, S. 2018.
\newblock Harmonious attention network for person re-identification.
\newblock In \emph{CVPR}, 2285--2294.

\bibitem[{Lin et~al.(2016)Lin, Dollár, Girshick, He, and
  Belongie}]{Lin2016Feature}
Lin, T.~Y.; Dollár, P.; Girshick, R.; He, K.; and Belongie, S. 2016.
\newblock Feature Pyramid Networks for Object Detection .

\bibitem[{Lin et~al.(2017{\natexlab{a}})Lin, Goyal, Girshick, He, and
  Dollar}]{Lin2017Focal}
Lin, T.~Y.; Goyal, P.; Girshick, R.; He, K.; and Dollar, P. 2017{\natexlab{a}}.
\newblock Focal Loss for Dense Object Detection.
\newblock \emph{TPAMI} PP(99): 2999--3007.

\bibitem[{Lin et~al.(2014)Lin, Maire, Belongie, Hays, Perona, Ramanan,
  Doll{\'a}r, and Zitnick}]{lin2014microsoft}
Lin, T.-Y.; Maire, M.; Belongie, S.; Hays, J.; Perona, P.; Ramanan, D.;
  Doll{\'a}r, P.; and Zitnick, C.~L. 2014.
\newblock Microsoft coco: Common objects in context.
\newblock In \emph{ECCV}, 740--755. Springer.

\bibitem[{Lin et~al.(2017{\natexlab{b}})Lin, Feng, Santos, Yu, Xiang, Zhou, and
  Bengio}]{lin2017structured}
Lin, Z.; Feng, M.; Santos, C. N.~d.; Yu, M.; Xiang, B.; Zhou, B.; and Bengio,
  Y. 2017{\natexlab{b}}.
\newblock A structured self-attentive sentence embedding.
\newblock \emph{arXiv preprint arXiv:1703.03130} .

\bibitem[{Linsley et~al.(2018)Linsley, Scheibler, Eberhardt, and
  Serre}]{linsley2018global}
Linsley, D.; Scheibler, D.; Eberhardt, S.; and Serre, T. 2018.
\newblock Global-and-local attention networks for visual recognition.
\newblock \emph{arXiv preprint arXiv:1805.08819} .

\bibitem[{Long, Shelhamer, and Darrell(2015)}]{long2015fully}
Long, J.; Shelhamer, E.; and Darrell, T. 2015.
\newblock Fully convolutional networks for semantic segmentation.
\newblock In \emph{CVPR}, 3431--3440.

\bibitem[{Mnih et~al.(2014)Mnih, Heess, Graves et~al.}]{mnih2014recurrent}
Mnih, V.; Heess, N.; Graves, A.; et~al. 2014.
\newblock Recurrent models of visual attention.
\newblock In \emph{NeurIPS}, 2204--2212.

\bibitem[{Newell, Yang, and Deng(2016)}]{newell2016stacked}
Newell, A.; Yang, K.; and Deng, J. 2016.
\newblock Stacked hourglass networks for human pose estimation.
\newblock In \emph{ECCV}, 483--499. Springer.

\bibitem[{Oliva et~al.(2003)Oliva, Torralba, Castelhano, and
  Henderson}]{Oliva2003et}
Oliva, A.; Torralba, A.; Castelhano, M.~S.; and Henderson, J.~M. 2003.
\newblock et al. Top-down control of visual attention in object detection.
\newblock \emph{ICIP} 5(8): 706--710.

\bibitem[{Pang et~al.(2020{\natexlab{a}})Pang, Li, Zhang, Li, and
  Lu}]{pang2020tubetk}
Pang, B.; Li, Y.; Zhang, Y.; Li, M.; and Lu, C. 2020{\natexlab{a}}.
\newblock TubeTK: Adopting Tubes to Track Multi-Object in a One-Step Training
  Model.
\newblock In \emph{CVPR}, 6308--6318.

\bibitem[{Pang et~al.(2019)Pang, Zha, Cao, Shi, and Lu}]{pang2019deep}
Pang, B.; Zha, K.; Cao, H.; Shi, C.; and Lu, C. 2019.
\newblock Deep RNN Framework for Visual Sequential Applications.
\newblock In \emph{CVPR}, 423--432.

\bibitem[{Pang et~al.(2020{\natexlab{b}})Pang, Zha, Cao, Tang, Yu, and
  Lu}]{pang2020complex}
Pang, B.; Zha, K.; Cao, H.; Tang, J.; Yu, M.; and Lu, C. 2020{\natexlab{b}}.
\newblock Complex sequential understanding through the awareness of spatial and
  temporal concepts.
\newblock \emph{Nature Machine Intelligence} 2(5): 245--253.

\bibitem[{Pang et~al.(2020{\natexlab{c}})Pang, Zha, Zhang, and
  Lu}]{pang2020adverb}
Pang, B.; Zha, K.; Zhang, Y.; and Lu, C. 2020{\natexlab{c}}.
\newblock Further Understanding Videos through Adverbs: {A} New Video Task.
\newblock In \emph{AAAI}, 11823--11830.

\bibitem[{Parmar et~al.(2019)Parmar, Ramachandran, Vaswani, Bello, Levskaya,
  and Shlens}]{parmar2019stand}
Parmar, N.; Ramachandran, P.; Vaswani, A.; Bello, I.; Levskaya, A.; and Shlens,
  J. 2019.
\newblock Stand-alone self-attention in vision models.
\newblock In \emph{NeurIPS}, 68--80.

\bibitem[{{Piao} et~al.(2019){Piao}, {Ji}, {Li}, {Zhang}, and
  {Lu}}]{Piao_2019_ICCV}
{Piao}, Y.; {Ji}, W.; {Li}, J.; {Zhang}, M.; and {Lu}, H. 2019.
\newblock Depth-induced Multi-scale Recurrent Attention Network for Saliency
  Detection.
\newblock In \emph{ICCV}.

\bibitem[{Redmon and Farhadi(2018)}]{redmon2018yolov3}
Redmon, J.; and Farhadi, A. 2018.
\newblock Yolov3: An incremental improvement.
\newblock \emph{arXiv preprint arXiv:1804.02767} .

\bibitem[{Ren and Zemel(2017)}]{ren2017end}
Ren, M.; and Zemel, R.~S. 2017.
\newblock End-to-end instance segmentation with recurrent attention.
\newblock In \emph{CVPR}, 6656--6664.

\bibitem[{Russakovsky et~al.(2015)Russakovsky, Deng, Su, Krause, Satheesh, Ma,
  Huang, Karpathy, Khosla, Bernstein et~al.}]{russakovsky2015imagenet}
Russakovsky, O.; Deng, J.; Su, H.; Krause, J.; Satheesh, S.; Ma, S.; Huang, Z.;
  Karpathy, A.; Khosla, A.; Bernstein, M.; et~al. 2015.
\newblock Imagenet large scale visual recognition challenge.
\newblock \emph{International journal of computer vision} 115(3): 211--252.

\bibitem[{Simonyan(2014)}]{simonyan2014very}
Simonyan, K. 2014.
\newblock Very deep convolutional networks for large-scale image recognition.
\newblock \emph{arXiv preprint arXiv:1409.1556} .

\bibitem[{Srivastava, Greff, and Schmidhuber(2015)}]{srivastava2015highway}
Srivastava, R.~K.; Greff, K.; and Schmidhuber, J. 2015.
\newblock Highway networks.
\newblock \emph{arXiv preprint arXiv:1505.00387} .

\bibitem[{Sun et~al.(2019)Sun, Xiao, Liu, and Wang}]{sun2019deep}
Sun, K.; Xiao, B.; Liu, D.; and Wang, J. 2019.
\newblock Deep high-resolution representation learning for human pose
  estimation.
\newblock \emph{arXiv preprint arXiv:1902.09212} .

\bibitem[{Tang et~al.(2020)Tang, Xia, Mu, Pang, and Lu}]{tang2020asynchronous}
Tang, J.; Xia, J.; Mu, X.; Pang, B.; and Lu, C. 2020.
\newblock Asynchronous Interaction Aggregation for Action Detection.
\newblock \emph{arXiv preprint arXiv:2004.07485} .

\bibitem[{Tian et~al.(2019)Tian, Shen, Chen, and He}]{tian2019fcos}
Tian, Z.; Shen, C.; Chen, H.; and He, T. 2019.
\newblock FCOS: Fully Convolutional One-Stage Object Detection.
\newblock \emph{arXiv preprint arXiv:1904.01355} .

\bibitem[{Vaswani et~al.(2017)Vaswani, Shazeer, Parmar, Uszkoreit, Jones,
  Gomez, Kaiser, and Polosukhin}]{vaswani2017attention}
Vaswani, A.; Shazeer, N.; Parmar, N.; Uszkoreit, J.; Jones, L.; Gomez, A.~N.;
  Kaiser, {\L}.; and Polosukhin, I. 2017.
\newblock Attention is all you need.
\newblock In \emph{NeurIPS}, 5998--6008.

\bibitem[{Wang et~al.(2015)Wang, Shen, Shao, Zhang, Xue, and
  Zhang}]{wang2015multiple}
Wang, D.; Shen, Z.; Shao, J.; Zhang, W.; Xue, X.; and Zhang, Z. 2015.
\newblock Multiple granularity descriptors for fine-grained categorization.
\newblock In \emph{ICCV}, 2399--2406.

\bibitem[{Wang et~al.(2017)Wang, Jiang, Qian, Yang, Li, Zhang, Wang, and
  Tang}]{wang2017residual}
Wang, F.; Jiang, M.; Qian, C.; Yang, S.; Li, C.; Zhang, H.; Wang, X.; and Tang,
  X. 2017.
\newblock Residual attention network for image classification.
\newblock In \emph{CVPR}, 3156--3164.

\bibitem[{Wang et~al.(2018)Wang, Girshick, Gupta, and He}]{wang2018non}
Wang, X.; Girshick, R.; Gupta, A.; and He, K. 2018.
\newblock Non-local neural networks.
\newblock In \emph{CVPR}, 7794--7803.

\bibitem[{Woo et~al.(2018)Woo, Park, Lee, and So~Kweon}]{woo2018cbam}
Woo, S.; Park, J.; Lee, J.-Y.; and So~Kweon, I. 2018.
\newblock Cbam: Convolutional block attention module.
\newblock In \emph{ECCV}, 3--19.

\bibitem[{Xiao, Wu, and Wei(2018)}]{xiao2018simple}
Xiao, B.; Wu, H.; and Wei, Y. 2018.
\newblock Simple baselines for human pose estimation and tracking.
\newblock In \emph{ECCV}, 466--481.

\bibitem[{Xie et~al.(2017)Xie, Girshick, Doll{\'a}r, Tu, and
  He}]{xie2017aggregated}
Xie, S.; Girshick, R.; Doll{\'a}r, P.; Tu, Z.; and He, K. 2017.
\newblock Aggregated residual transformations for deep neural networks.
\newblock In \emph{CVPR}, 1492--1500.

\bibitem[{Xiu et~al.(2018)Xiu, Li, Wang, Fang, and Lu}]{xiu2018pose}
Xiu, Y.; Li, J.; Wang, H.; Fang, Y.; and Lu, C. 2018.
\newblock Pose flow: Efficient online pose tracking.
\newblock \emph{arXiv preprint arXiv:1802.00977} .

\bibitem[{Xu et~al.(2015)Xu, Ba, Kiros, Cho, Courville, Salakhudinov, Zemel,
  and Bengio}]{xu2015show}
Xu, K.; Ba, J.; Kiros, R.; Cho, K.; Courville, A.; Salakhudinov, R.; Zemel, R.;
  and Bengio, Y. 2015.
\newblock Show, attend and tell: Neural image caption generation with visual
  attention.
\newblock In \emph{ICML}, 2048--2057.

\bibitem[{Yu et~al.(2018)Yu, Wang, Shelhamer, and Darrell}]{YuDeep}
Yu, F.; Wang, D.; Shelhamer, E.; and Darrell, T. 2018.
\newblock Deep Layer Aggregation.
\newblock In \emph{CVPR}.

\bibitem[{Yue and et~al.(2018)}]{KaiyuCompact}
Yue, K.; and et~al. 2018.
\newblock Compact Generalized Non-local Network.
\newblock In \emph{NeurIPS}.

\bibitem[{Zhang et~al.(2018)Zhang, Li, Li, Wang, Zhong, and
  Fu}]{zhang2018image}
Zhang, Y.; Li, K.; Li, K.; Wang, L.; Zhong, B.; and Fu, Y. 2018.
\newblock Image super-resolution using very deep residual channel attention
  networks.
\newblock In \emph{ECCV}, 286--301.

\bibitem[{Zhou et~al.(2018)Zhou, Andonian, Oliva, and Torralba}]{ZhouTemporal}
Zhou, B.; Andonian, A.; Oliva, A.; and Torralba, A. 2018.
\newblock Temporal Relational Reasoning in Videos.
\newblock In \emph{ECCV}.

\bibitem[{Zhu et~al.(2019)Zhu, Cheng, Zhang, Lin, and Dai}]{zhu2019empirical}
Zhu, X.; Cheng, D.; Zhang, Z.; Lin, S.; and Dai, J. 2019.
\newblock An empirical study of spatial attention mechanisms in deep networks.
\newblock \emph{arXiv preprint arXiv:1904.05873} .

\end{thebibliography}
	}
	
\end{document}